\documentclass[a4paper,conference]{IEEEtran}
\usepackage[utf8]{inputenc}
\usepackage[pdftex]{graphicx}
\usepackage{array}
\usepackage{mdwmath}
\usepackage{mdwtab}
\usepackage{amssymb,latexsym}
\usepackage{stfloats}
\usepackage{amsmath}
\usepackage{subfig}
\usepackage{url}
\usepackage{algcompatible}

\usepackage{algorithm}

\usepackage[font={footnotesize}]{caption}

\usepackage{balance}

\usepackage[T1]{fontenc}

\usepackage[compatible]{algpseudocode}

\makeatletter
\def\mycopyrightnotice{\footnotesize 979-8-3503-7542-8/24/\$31.00 \copyright2024 IEEE\hfill}
\def\ps@IEEEtitlepagestyle{%
  \def\@oddfoot{\mycopyrightnotice}%
  \def\@evenfoot{}%
}

\begin{document}

\title{Enhanced Drift-Aware Computer Vision Architecture for Autonomous Driving}

\author{\IEEEauthorblockN{1\textsuperscript{st} Md Shahi Amran Hossain}
\IEEEauthorblockA{\textit{Chair of Embedded Systems} \\
\textit{University of Siegen}\\
Siegen, Germany \\
ORCID: 0009-0000-9466-0912}
\and
\IEEEauthorblockN{2\textsuperscript{nd} Abu Shad Ahammed }
\IEEEauthorblockA{\textit{Chair of Embedded Systems} \\
\textit{University of Siegen}\\
Siegen, Germany \\
ORCID: 0009-0007-6715-8098}
\and
\IEEEauthorblockN{3\textsuperscript{rd} Sayeri Mukherjee}
\IEEEauthorblockA{\textit{Chair of Embedded Systems} \\
\textit{University of Siegen}\\
Siegen, Germany \\
ORCID: 0009-0008-6533-9538
}
\and

\IEEEauthorblockN{4\textsuperscript{th} Roman Obermaisser}
\IEEEauthorblockA{\textit{Chair of Embedded Systems} \\
\textit{University of Siegen}\\
Siegen, Germany \\
ORCID: 0009-0002-4483-1503}
}
\maketitle

% As a general rule, do not put math, special symbols or citations
% in the abstract or keywords.
\begin{abstract}
The use of computer vision in automotive is a trending research in which safety and security are a primary concern. In particular, for autonomous driving, preventing road accidents requires highly accurate object detection under diverse conditions. To address this issue, recently the International Organization for Standardization (ISO) released the 8800 norm, providing structured frameworks for managing associated AI relevant risks. However, challenging scenarios such as adverse weather or low lighting often introduce data drift, leading to degraded model performance and potential safety violations. In this work, we present a novel hybrid computer vision architecture trained with thousands of synthetic image data from the road environment to improve robustness in unseen drifted environments. Our dual mode framework utilized YOLO version 8 for swift detection and incorporated a five-layer CNN for verification. The system functioned in sequence and improved the detection accuracy by more than 90\% when tested with drift-augmented road images. The focus was to demonstrate how such a hybrid model can provide better road safety when working together in a hybrid structure.
\end{abstract}

% Note that keywords are not normally used for peerreview papers.

% For peer review papers, you can put extra information on the cover
% page as needed:
% \ifCLASSOPTIONpeerreview
% \begin{center} \bfseries EDICS Category: 3-BBND \end{center}
% \fi
%
% For peerreview papers, this IEEEtran command inserts a page break and
% creates the second title. It will be ignored for other modes.
\IEEEpeerreviewmaketitle

\section{Introduction}
%Sayeri
Autonomous vehicles are rapidly becoming a reality and reshaping our transportation methods. These smart cars can navigate complex environments with impressive accuracy, handling essential tasks such as lane detection, obstacle avoidance, and real-time navigation, all with minimal human input. This technological evolution is largely driven by advances in artificial intelligence (AI) algorithms, which enable vehicles to interpret and respond to their surroundings effectively. Computer vision (CV), a part of AI, is a trending technology now that works on vision-based perception systems and has the ability to detect objects on the driving path, understand scenarios and assist drivers in navigation, ultimately promising to improve road safety and traffic efficiency \cite{wen2022deep}.\\
As we move toward the widespread deployment of autonomous vehicles, we face substantial challenges, particularly in ensuring their safety and reliability. These vehicles must not only navigate complex environments, but also do so with precision in unpredictable real-world conditions. To address these challenges, our study introduces an innovative hybrid model framework that combines two advanced computer vision techniques working together to improve the safety and robustness of autonomous vehicles. The developed framework contains YOLO and a convolutional neural network (CNN) architecture, both trained with synthetic datasets generated from the open-source CARLA automotive simulator. Generally, such CV models are expected to work in normal road environments, but may have trouble if there exist abnormal environments such as changes in lighting and weather, failure in camera sensor, etc. To address these challenges, we augmented our environmental conditions and prepared a dataset that simulates drift and examined the resilience and stability of the framework. This dual model framework is developed and optimized with the expectation that they will have the same classification result when detecting objects. But in instances where the two models produce conflicting outputs, indicative of potential errors or uncertainty, the system promptly transitions to a predefined safe state, prioritizing safety by mitigating risks under ambiguous conditions.\\
The remainder of this manuscript is organized as follows. Section II reviews CV approaches in the automotive domain. Section III addresses data drift and mitigation strategies. Section IV outlines the CV models used, while Section V discusses synthetic data generation to develop drift-aware models. Section VI covers data organization for object and drift detection. Section VII presents the hybrid model pipeline. Section VIII details the real-time, drift-aware detection framework. Section IX evaluates system performance, and Section X concludes with key findings and future directions.

\section{Background and Literature Review}
Exploring previous research works relevant to this field, we found that many researchers are exploring the possibility of using computer vision models for road safety assessment \cite{dixit2021safety}, accurate localization to prevent possible deterioration of navigation performance\cite{chuprov2023robust}, or control vehicles to avoid accidents \cite{kumer2021controlling}. While prior research has explained the significance of computer vision in enhancing safe transportation, it has not vastly explored the vulnerabilities arising from false detections in real-time drift scenarios. The novelty of our research lies not only in the application of CV within the mobility domain but also in the development of a hybrid model designed to operate during automotive runtime, with the goal of improving system robustness and security under drift conditions.\\
Ensuring safe transportation is a critical objective in modern mobility systems, particularly with the rise of autonomous and semi-autonomous vehicles. Koopman and Wagner in their research \cite{koopman2017autonomous} discussed that the safe operation of autonomous vehicles requires a deep understanding of safety engineering, reliable system design, and the ability to address the complexity of dynamic real-world conditions through robust safety mechanisms. When there is a drift condition, the computer vision model integrated in the autonomous vehicle becomes vulnerable and can lead to a serious accident. The authors of this paper in their previous research work \cite{hossain2024impact, ahammed2025computer} explained the significant impact of drift scenarios on the reliability of safety critical systems and why they should be addressed through adaptive techniques that respond to changing operational conditions.\\
There are different computer vision models used in trending research of autonomous vehicles, including CNN, YOLO, CenterNet / CornerNet, Waymo’s VectorNet for object detection. Among these models, CNNs are widely used in object detection due to their effectiveness in feature extraction and classification \cite{garg2023review}. Although these models are powerful and offer good accuracy, they often require optimization for deployment in real-time applications like autonomous driving. To meet real-time constraints, YOLO (You Only Look Once) has emerged as a preferred detection framework due to its fast single-pass architecture. Redmon et al. \cite{redmon2016you} introduced YOLO as a unified model that performs detection in real time without significantly compromising accuracy, making it suitable for embedded autonomous driving systems. However, its performance can vary under unseen or drifted conditions, where CNN models can excel as they are capable of analyzing features with finer details to detect smaller or even overlapping objects \cite{neha2025classical, drid2020object}.\\
The reviews above made it clear that both models have their own advantages and disadvantages. Our research aims to combine them to a hybrid structure and verify if it provides better accuracy in the automotive environment. A similar approach has been offered by Viswanatha et al. \cite{cr2022real} which addresses the integration of these models in industries like finance, emphasizing their role in feature extraction and target detection. Their findings support the idea that such hybrid systems can improve detection accuracy by mitigating false positives and missed detections, especially in drift scenarios.
\section{Impact of Data Drift and Mitigation Pathway}
\begin{figure}[!b]
\center
\subfloat[Tilted Road Sign in Low Light]{\label{fig:drift1}
\includegraphics[width=0.24\textwidth]{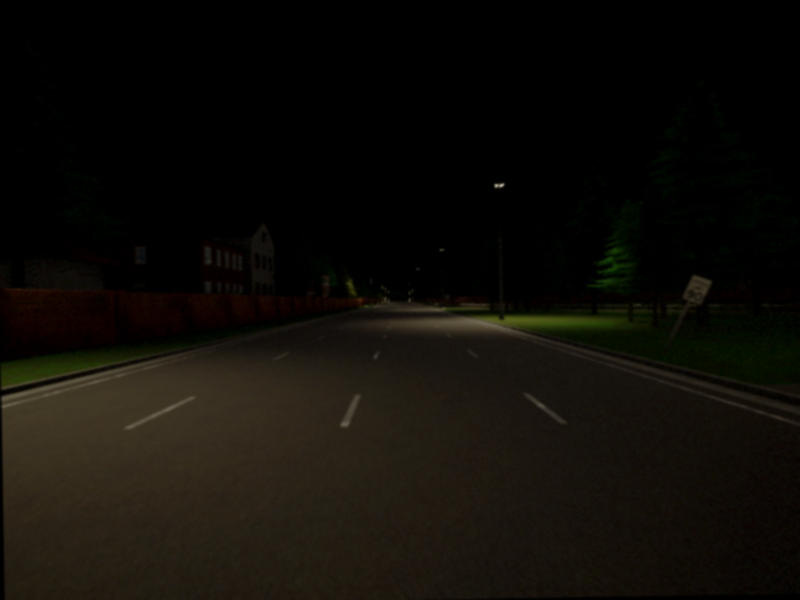}}
\subfloat[High Sunlight Reflection]{\label{fig:drift2}
\includegraphics[width=0.24\textwidth]{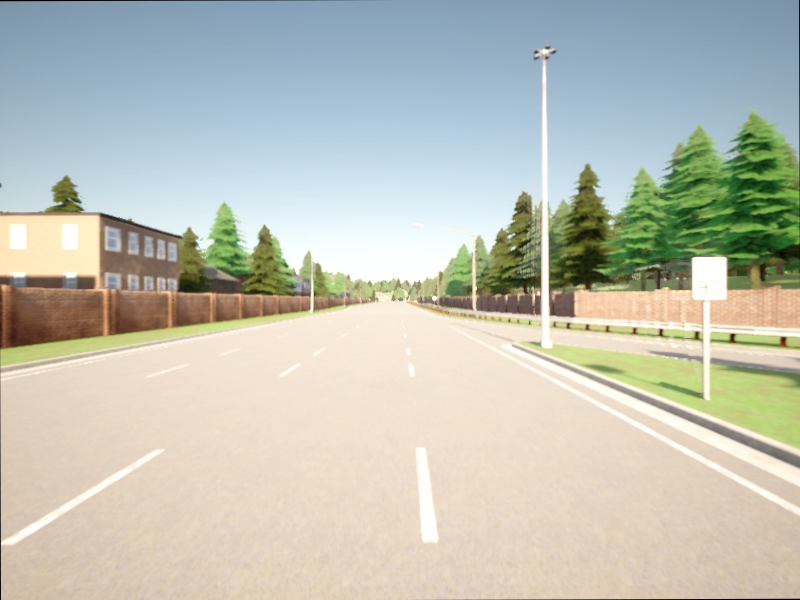}}
\caption{Example of Data Drift in Synthetic Images of Road Environment}
\label{fig:drift}
\end{figure}

The adverse impact of data drift in CV-based autonomous driving is currently a major concern where drift scenarios can affect the lives of road users, as well as cause environmental and economic damage. It implies changes in the statistical properties and characteristics over time of the input domain which may have a significant effect on the performance of the computer vision model. In our previous research \cite{hossain2024impact}, we found that the YOLO model had a degraded accuracy by almost 40\% when drift scenarios are tested against normal scenarios. Our solution was to augment the model training data to address drift, which resulted in higher accuracy. However, a major concern in this approach was that the drift that we assumed and integrated into our data set cannot guarantee robust real-world performance, while there can be different drifts faced during autonomous driving, which our drift-train model may not detect. Considering this limitation, we proposed a hybrid model framework with YOLO and CNN which can tackle any trained or untrained drift situation when detecting objects. In this framework, YOLO is utilized as a real-time object detector, efficiently identifying and localizing objects within complex driving environments. CNN model will work subsequently for a fine-grained recognition of that object and classify it. By combining the quick detection ability of YOLO with the accurate classification strength of CNN, the hybrid architecture offers a synergistic model that improves detection accuracy and increases interpretability. Figure \ref{fig:drift} demonstrates instances of data drift occurring due to variations in lighting conditions and the placement of road signs.
\section{Computer Vision Models: YOLO and CNN}
The YOLO (You Only Look Once) model is an open-source family of object detection models that has made a significant breakthrough in the field of computer vision. One of the core capabilities of YOLO over other object detection models such as the single shot multi-box detector (SSD), Region-based CNN is its high speed and accuracy, enabling real-time processing, which is critical in applications ranging from autonomous vehicles to surveillance systems \cite{T_Liaqat_2024}. In this study, one of the recent versions of the YOLO family YOLOv8 is used over YOLOv9 and YOLOv10 due to its stability and precision efficiency trade-off \cite{C_Silva_2025}.\\ 
Convolutional Neural Networks (CNNs) have emerged as a powerful tool in the realm of deep learning, especially for image recognition and classification tasks. Therefore, CNN is used parallelly to classify road signs and is applied only to the segmented part of the images using the YOLO model. The CNN model we developed consists of several individual layers as described below.
\begin{itemize}
    \item \textbf{Convolutional layer}: This layer is the fundamental component of a CNN model and extracts the feature from the input, such as the edge, texture, and shape. This is achieved by utilizing filters or kernels that slide through the whole input according to the stride.
    \item \textbf{Pooling layer}: The pooling layer usually follows a convolution layer, and this layer is used primarily to reduce the spatial dimensions of the feature maps, thus decreasing the computational load and helping to mitigate overfitting. These pooling techniques summarize features in a specific region of the feature map \cite{C_Asif_2022}.
    \item \textbf{Fully connected layer}: These layers are located at the end of the CNN model architecture, consist of neurons that are fully connected to the neurons of the next layer. This setup enables the integration of features learned in preceding layers and helps make the ultimate classification decision. In the final fully Connected Layer, we used the softmax function to get the classification result.
\end{itemize}
\section{Synthetic Data Generation}.
The use of synthetic data for CV models has a few key advantages, such as simplicity and often less time consuming, enabling the creation of different scenarios and avoiding privacy issues. But this process also has its drawbacks, which is the quality of data with respect to the real world. On the other side, automotive data containing drift environment is difficult to get as no such repositories are currently available. Furthermore, there are challenges associated with the acquisition of real-world data in autonomous driving research, such as safety risks, cost constraints, and limited data diversity. Considering these aspects, we have used the open source car learning-to-act (CARLA) automotive simulator to generate synthetic photorealistic data. The simulator provides a flexible and controlled environment for simulating complex driving scenarios, which is essential for training and validating machine learning models in the automotive domain \cite{ly2020learning}.\\
The CARLA simulator features a diverse set of urban and rural environments, various vehicle models, and realistic sensory data \cite{L_Rong_2020}. It also allows for the creation of various scenarios, including those that are difficult or dangerous to replicate in real life, such as extreme weather conditions or complex traffic situations \cite{C_Jang_2021}. The ability to simulate different weather conditions, times of day, and traffic scenarios enhances the robustness of the data generated, making it suitable for real-world applications \cite{A_Gutirrez_2020}. Synthetic images were collected from the CARLA simulator through the Robot Operating System (ROS), and the CARLA-ROS bridge was used to establish the connection between CARLA and ROS. The collection of image data occurred in two phases. Initially, images depicting typical road settings were collected from various urban and rural maps to create a baseline image dataset. Subsequently, in the second phase, we added drift to the data set by changing the shape, adjusting daylight, and inputting Gaussian noise. Seven different objects were placed on the maps for detection which are: Round-shaped 30, 60 and 90 kilometers speed limit; Square shaped 30, 60 and 90 kilometers speed limit and Stop sign.
\section{Data Organization and Management}
The CARLA simulator allows the production of the self-labeled image, which reduces the number of manual annotations and provides rapid iteration in the development of ML models \cite{D_Komorniczak_2022}. However, manual labeling remains a critical aspect of synthetic data generation, particularly when it comes to ensuring the quality and accuracy of the generated datasets. It often facilitates the achievement of high fidelity in the dataset, particularly in complex scenarios where nuanced understanding is required \cite{A_Bemposta_2023}. Therefore, manual labeling was performed after collecting the images from the CARLA simulator, which were carefully inspected and filtered. After filtering was completed, we selected 2,017 images without drift and 600 images with drift. An open source web-based tool called CVAT (Computer Vision Annotation program), created by Microsoft, was used to annotate the images as a requirement for the YOLO model training. This tool ensured accuracy in labeling images with appropriate class labels, which significantly contributes to model performance. After the labeling is completed, the standard dataset for the object detection algorithm is divided into three parts where 80 percent of the data is assigned to the training, 10 percent to the test and 10 percent to the validation subset. This division is done to maintain a balanced evaluation framework. However, the drift data set is completely unchanged and is used only for testing purposes to ensure the effectiveness of the models in detecting distributional shifts.\\
\section{Development of the Hybrid Model}
The development of the hybrid model begins with training of the YOLO model after performing the necessary pre-processing and organization of the synthetic data generated by CARLA. All acquired data was placed in a centralized folder structure according to the YOLO requirement. Once the classes and data storage links are provided through a '.yaml' file, the model architecture goes through the training phase. This configuration includes selecting the backbone network, essential for feature extraction, and setting several hyperparameters such as learning rate, batch size, and the number of epochs. After completion of the training process, the performance of the YOLO model was evaluated using standard metrics such as precision, recall, F1 score, and mAP.\\
The next step was to train CNN model, which takes a different way as it has no capability of doing object localization, rather it requires to have the image of an object. Therefore, it became necessary to create a new data set based solely on objects, achieved by manually cropping the original YOLO training data with a Python script. Annotated images through CVAT were used to determine the location and size of objects in each image with center position (x,y) and dimensions (w,h), which are normalized values relative to the width and height of the image. A bounding-box cropping method was adopted to transform the YOLO-labeled object detection data set into a format suitable for the Convolutional Neural Network (CNN). This method can be divided into the following steps:
\begin{itemize}
    \item \textbf{Annotation analysis}: Each YOLO annotation file was processed. The normalized coordinates of the bounding box for each object in the images are converted to pixel values with the following equations:
    \[
x_1 = \left(x - \frac{w}{2}\right) \cdot \text{width}, \quad y_1 = \left(y - \frac{h}{2}\right) \cdot 
\text{height}
\]
\[
x_2 = \left(x + \frac{w}{2}\right) \cdot \text{width}, \quad y_2 = \left(y + \frac{h}{2}\right) \cdot \text{height}
\]
    \item \textbf{Bounding Box Extraction}: The boundary boxes were used to extract object-specific regions from the images. Each region of interest (ROI) was cropped and isolated.
    \item \textbf{Resizing}: The CNN model requires only one size of data as input. So each cropped image is resized to a fixed size of 64×64 pixels and the three RGB channels are retained.
    \item \textbf{Dataset Structure}: Each cropped image is stored separately, and its label is determined from the YOLO annotations.
    \item \textbf{Quality Check}: Data was checked manually to ensure that cropping and labeling were performed correctly.
\end{itemize}
 In our research, the custom CNN model developed consists of 3 convolution layers, 2 max-pool layers, and 2 dense layers.
\section{Real-Time Drift-Aware Object Detection}
\label{sec:driftaware}
\begin{figure}[!b]
\center
\includegraphics[width=0.48
\textwidth]{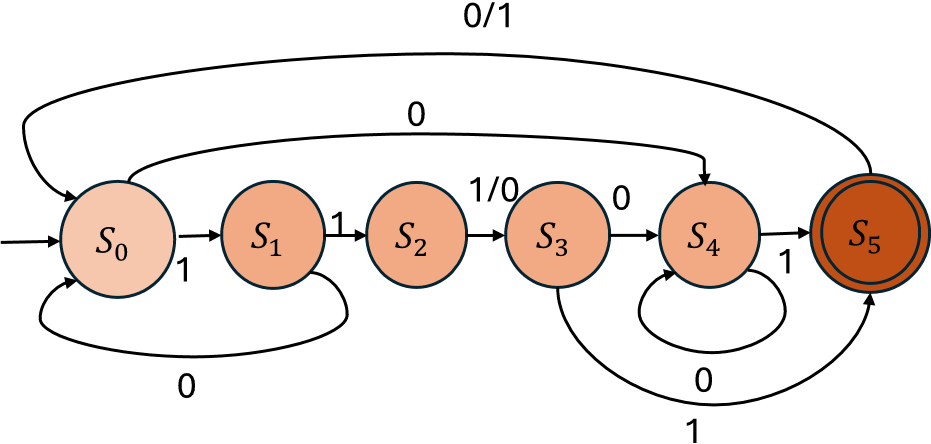}
\caption{State Machine of the Drift-aware Object Detection Framework}
\label{fig:state_machine} 
\end{figure}

The object detection system introduced in this work is designed to identify objects under various drift conditions in real time and assess when model updates are necessary after a detection failure. Instead of relying exclusively on the YOLO model, we build up the hybrid architecture adding a separate CNN model which was trained on the same training data set used with YOLO for object classification, and the conclusion was reached by evaluating the votes from the predictions of both models. The entire framework is mathematically modeled as a deterministic finite automaton (DFA) with five tuples and can be defined as follows:\\
\[
M = (Q, \Sigma, \delta, q_0, F)
\]
Where,
\begin{itemize}
    \item $Q$:  Finite set of states.
    \item $\Sigma$: A finite set of input symbols (alphabet).
    \item $\delta$: The transition function, $\delta : Q \times \Sigma \to Q$, mapping a state and an input symbol to a new state.
    \item $q_0$: The initial state, $q_0 \in Q$.
    \item $F$: The set of final states, $F \subseteq Q$.
\end{itemize}
Formal definition of the designed DFA according to the tuples,
\begin{itemize}
    \item \textbf{States:} $Q = \{S_0, S_1, S_2, S_3, S_4, S_5\}$ \\
          Where:
          \begin{itemize}
              \item $S_0$: Initial and data collection state
              \item $S_1$: Object detection and primary classification state
              \item $S_2$: Secondary classification state
              \item $S_3$: Decision making state
              \item $S_4$: Safe state
              \item $S_5$: Final state
          \end{itemize}
    \item \textbf{Inputs:} $\Sigma = \{0, 1\}$
    \item \textbf{Transition Function:} $\delta$ for each state is defined as:
          \[
          \begin{aligned}
              \delta(S_0) &= \{(0, S_4), (1, S_1)\} \\
              \delta(S_1) &= \{(0, S_0), (1, S_2)\} \\
              \delta(S_2) &= \{(0, S_3), (1, S_3)\} \\
              \delta(S_3) &= \{(0, S_4), (1, S_5)\} \\
              \delta(S_4) &= \{(0, S_4), (1, S_5)\} \\
              \delta(S_5) &= \{(0, S_0), (1, S_0)\}
          \end{aligned}
          \]
    \item \textbf{Initial State:} $q_0 = S_0$
    \item \textbf{Final State:} $F = \{S_5\}$
\end{itemize}
This state machine describes the entire detection process of the proposed pipeline in the automotive runtime. It takes two types of input (0 and 1) with six states; each must complete certain tasks to move to another state. The task for each state can be described as below.
\begin{itemize}
    \item Initial and data collection state {$S_0$}: This is the first state, as well as the image data collection state. If the image collection process from the RGB camera sensor is completed successfully (input 1), then this state goes to state $S_1$, otherwise (input 0) goes to state $S_4$.
    \item  Object detection and primary classification state {$S_1$}: In this state, the object detection algorithm (YOLO) is applied to the image collected in the state $S_0$. If YOLO detects on the trained objects (input 1), then the classification result and the segmented object of the image will be temporarily stored in this state. In contrast, if no object is detected (input 0), the system transitions to the state $S_0$.
    \item Secondary classification state {$S_2$}: The CNN model is applied in this state on the segmented image from state $S_1$, which is completely isolated from the primary model. After obtaining the classification result, whether it is 0 or 1, it moves to state $S_3$.
    \item Decision-making state $S_3$: The result of the YOLO and the CNN model is compared in this state and the decision is made upon the voting result of these two models. If both models predict the same class(input 1) then the system goes to the final state $S_5$. However, if the prediction between the two models differs (input 0), then it results in the state $S_4$ and it also indicates that there is a significant drift in the collected scenario, which poses a threat to the autonomous driving environment.
    \item Safe state {$S_4$}: If the system enters into this safe state, it will stay in this state(input 0) until the system receives any reset signal (input 1). If the system gets a reset signal (input 1), then it moves to the final state $S_5$.
    \item Final state {$S_5$}: For any given valid input (1 and 0), the system processes to the initial and data collection state $S_0$ and the system is ready to take a new input image.
\end{itemize}
In Figure \ref{fig:state_machine}, a visual demonstration of the state machine pipeline is provided. 
\section{Results and Discussions}
\begin{figure}[!b]
\center
\includegraphics[width=1\linewidth]{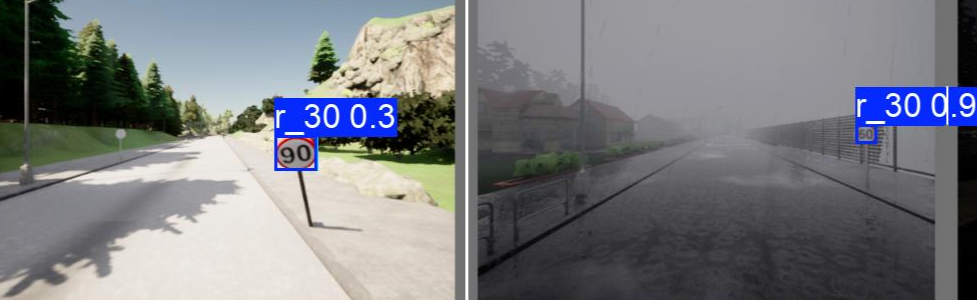}
\caption{Performance of YOLO model in Drift Situations. Original road signs were for 90 km/h and 60 km/h}.
\label{fig:wrongpredict} 
\end{figure}

\begin{figure}[!t]
\center
\includegraphics[width=1\linewidth]{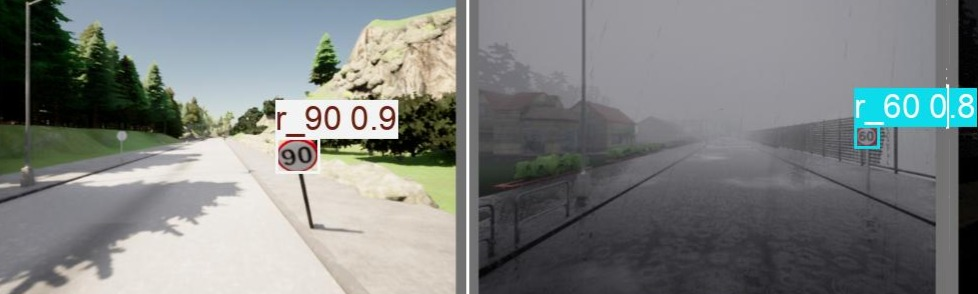}
\caption{Performance of Hybrid model in Drift Situations}
\label{fig:rightpredict} 
\end{figure}

Evaluation of our research work is done in three steps. First, we measure the performance of the YOLO model using both standard test data and drift data. Then, we evaluate the CNN model in the same way. In the last step, we integrate both models to create the hybrid system and assess whether it improves performance. This ensures that, if implemented in a real driving environment, the model can also manage unforeseen drift. In Table \ref{tab:yolo_test} and Table \ref{tab:cnn_test}, the performance of the individual model is depicted.\\
\begin{table}[!h]
\renewcommand{\arraystretch}{1.3}
\caption{\textsc{Performance of YOLO Model}}
\label{tab:yolo_test}
\centering
\begin{tabular}{|l|c|c|}
\hline
\textbf{Criteria} & \textbf{Standard Dataset} & \textbf{Drift Dataset} \\ \hline
Accuracy & 0.9901 & 0.8133 \\ \hline
Precision & 0.9897 & 0.8301  \\ \hline
Recall & 0.9899 & 0.8159  \\ \hline
F1-Score & 0.9895 & 0.8161  \\ \hline
\end{tabular}
\end{table}
\begin{table}[!h]
\renewcommand{\arraystretch}{1.3}
\caption{\textsc{Performance of CNN Model}}
\label{tab:cnn_test}
\centering
\begin{tabular}{|l|c|c|}
\hline
\textbf{Criteria} & \textbf{Standard Dataset} & \textbf{Drift Dataset} \\ \hline
Accuracy & 0.9938 & 0.9516 \\ \hline
Precision &  0.9922 & 0.9515 \\ \hline
Recall &  0.9875 &  0.9520  \\ \hline
F1-Score & 0.9894 & 0.9517  \\ \hline
\end{tabular}
\end{table}
On the standard dataset, the YOLO model exhibited high performance, achieving more than 95\% in each metric. It indicates that the model was capable of highly accurate object detection in ideal and well-represented environmental conditions. However, when evaluated on the drift dataset, performance was significantly degraded. The accuracy dropped to 0.8133, while the precision and recall decreased to 0.8301 and 0.8159, respectively. This performance degradation highlighted a key limitation of the implementation of a standalone YOLO model in safety-critical applications of autonomous driving, where the reliability of object detection is highly important because any misclassification or missed detection can cause serious accidents. Figure \ref{fig:wrongpredict} illustrates an instance in which the YOLO model inaccurately predicts the speed limit of road signs in the presence of drift scenarios.\\
The CNN model demonstrated better performance compared to the YOLO model, particularly in the drift data set, which are critical in the context of safety-critical automotive systems. As shown in Table~\ref{tab:cnn_test}, the CNN achieved an accuracy of 0.9938 on the standard data set, with precision, recall, and F1 score values of 0.9922, 0.9875, and 0.9894, respectively. These metrics indicate that the model was highly effective in correctly identifying and classifying objects in a normal driving environment. However, more importantly, within the drift dataset, which was intended to simulate real-world interruptions, CNN maintained strong performance. The accuracy only slightly decreased to 0.9516, and the precision, recall, and F1 score values remained consistently high at 0.9515, 0.9520, and 0.9517, respectively. The final step was performed by testing the hybrid architecture only with the 600 drift data, and the performance overview can be seen in Table \ref{tab:hybridperf}.
\begin{table}[!b]
\renewcommand{\arraystretch}{1.3}
\caption{\textsc{Performance of Hybrid Model}}
\label{tab:hybridperf}
\centering
\begin{tabular}{|l|c|}
\hline
\textbf{Criteria} & \textbf{Drift Dataset} \\ \hline
Accuracy & 0.9716 \\ \hline
Precision &  0.9958 \\ \hline
Recall &  0.9696 \\ \hline
F1-Score & 0.9819 \\ \hline
\end{tabular}
\end{table}
The hybrid model architecture demonstrated robust and improved performance under drift conditions compared to the standalone CNN. It achieves an accuracy of 0.9716, reflecting a substantial level of precision across all categories. In particular, the precision was 0.9958, which indicated that nearly every positive prediction was correct, with very few false positives. Meanwhile, the recall was 0.9696, highlighting the effectiveness of the model in accurately identifying most real positive cases, with minimal false negatives. Figure \ref{fig:rightpredict} illustrates an example of the high accuracy performance of the hybrid model in drift scenarios where the YOLO model failed, as shown in Figure \ref{fig:wrongpredict}.
\\
The YOLO model exhibited the weakest performance on drift data, correctly classifying only 488 out of 600 samples, resulting in 112 false detections, each of which can pose a serious risk in safety-critical autonomous driving applications. The CNN model improved considerably, achieving 571 correct predictions and reducing the total number of misclassifications to 29. However, the most reliable performance was achieved using the hybrid framework, which produced 583 correct predictions, closely matching CNN in overall accuracy, but with far greater emphasis on safety. Specifically, the hybrid model successfully identified 105 cases that were missed by the YOLO model, missing only 15 actual threats detected and classified correctly by YOLO. In all those 120 cases, the hybrid architecture will raise an alarm, indicating there are high drift scenarios ahead. Only two cases were missed by both models, which is a 98.21\% reduction in hazardous oversights compared to YOLO. It confirms that the hybrid structure significantly improves the reliability of real-time object detection under domain drift and is better suited for deployment in safety-critical autonomous systems.\\
In real-world driving scenarios, rapid inference is a critical requirement for computer vision models integrated into safety-critical autonomous systems. Among the models evaluated, the YOLO architecture demonstrated the fastest inference time, averaging 2.4 milliseconds, making it highly suitable for time-sensitive applications. In comparison, the CNN model achieved an inference time of 28.34 milliseconds, while the proposed hybrid framework maintained a comparable latency of 30.74 milliseconds. Although the hybrid approach incurs slightly higher computational cost, it remains within an acceptable real-time operational range, offering a favorable trade-off between detection robustness and response time under drift.
\section{Conclusion}
Our enhanced hybrid computer vision approach provided better performance in terms of objects, like road sign detection, than individual models. Instead of training the model with specific drift conditions, we provided a solution that can handle unknown drifts as well. This model can be implemented in the future in embedded architectures of safety critical automotive systems and tested with real-world automotive scenarios.
\section*{Acknowledgment}
This work has been funded by 1) the Federal Ministry of Education and Research (BMBF) as part of AutoDevSafeOps (01IS22087Q) and 2) the research project EcoMobility through the European Commission (101112306) and BMBF (16MEE0316).
\bibliographystyle{ieeetr}
\bibliography{main}

\end{document}